\documentclass[a4paper]{article}

\usepackage{cite}
\usepackage{INTERSPEECH2021}
\usepackage{amsmath,graphicx}
\usepackage{xcolor}
\usepackage{amssymb,amsmath,epsfig,cite,url,multicol,multirow}
\usepackage{hyperref}
\usepackage{graphicx}
\usepackage{float}
\usepackage{comment}
\usepackage{float}
\usepackage{diagbox}
\floatstyle{plaintop}
\restylefloat{table}

\title{Knowledge Distillation from BERT Transformer to Speech Transformer for Intent Classification}
\name{Yidi Jiang, Bidisha Sharma, Maulik Madhavi, and  Haizhou Li}
\address{
Department of Electrical and Computer Engineering,\\
National University of Singapore, Singapore
}
\email{E0572589@u.nus.edu,\{s.bidisha,maulik.madhavi, haizhou.li\}@nus.edu.sg}
\begin{document}
\maketitle
%
\begin{abstract}
End-to-end intent classification using speech has numerous advantages compared to the conventional pipeline approach using automatic speech recognition (ASR), followed by natural language processing modules. It attempts to predict intent from speech without using an intermediate ASR module. However, such end-to-end framework suffers from the unavailability of large speech resources with higher acoustic variation in spoken language understanding. In this work, we exploit the scope of the transformer distillation method that is specifically designed for knowledge distillation from a transformer based language model to a transformer based speech model. In this regard, we leverage the reliable and widely used bidirectional encoder representations from transformers (BERT) model as a language model and transfer the knowledge to build an acoustic model for intent classification using the speech. In particular, a multilevel transformer based teacher-student model is designed, and knowledge distillation is performed across attention and hidden sub-layers of different transformer layers of the student and teacher models. We achieve an intent classification accuracy of 99.10\% and 88.79\% for  Fluent speech corpus and ATIS database, respectively. Further, the proposed method demonstrates better performance and robustness in acoustically degraded condition compared to the baseline method.

\end{abstract}

\vspace{-0.2cm}
\section{Introduction}
The spoken language understanding (SLU) systems aim to derive the meaning of users' queries, which helps the conversational agent to respond appropriately to the user. The SLU performs several tasks, namely, domain identification, intent classification, and slot filling. As a crucial part of SLU, the intent classification module is designed to extract the meaning or intention of the user's spoken utterance. There is a wide range of interactive interface-based applications requiring SLU systems, namely, voice search and chatbot on mobile devices, attracting interest from both commercial and academic sectors.

The conventional pipeline SLU system for intent classification contains two modules, namely, an automatic speech recognition (ASR) system that converts the speech to text sequence, and a natural language understanding (NLU) module that maps the text transcript to the speaker’s intent. The end-to-end architectures are widely used in several domains, like ASR \cite{ref7_deepspeech2,ref8_ChanJLV16,ref9_neuralasr}, speech synthesis \cite{ref10_OordDZSVGKSK16}, machine translation \cite{ref11_SutskeverVL14,ref12_BahdanauCB14,ref13_CNNseq2ses}. The end-to-end frameworks can be used to overcome the issues with the conventional pipeline SLU system,  which has also gained popularity in the field of SLU~\cite{serdyuk2018towards,price2020improved, haghani2018audio}.

The end-to-end SLU acts as an individual single model and it directly predicts the intent from speech without exploiting an intermediate text representation from ASR. There are many advantages of using the end-to-end system against the conventional pipeline SLU system. In particular, it aims to optimize the performance metric which is intent classification accuracy directly without using any intermediate ASR modeling. This results into a compact neural network model and eliminates the issue caused by intermediate transcription error \cite{ref1_lugosch_speech_2019}.  
Secondly, human interpretation of a spoken utterance depends on prosody, speech rate, and loudness, which are not considered in the pipeline SLU system.

Due to the highly variable and complex nature of speech, a large amount of data and intensive computational resources are required for the training of end-to-end system. The computer vision \cite{ref21_YosinskiCBL14,ref22_SimonyanZ14a,zhou2020unified}, natural language processing (NLP) \cite{ref23_DaiL15,ref24_RuderH18,ref25_radford2018improving,ref26_SanhWR19,ref27_bert_paper}, and ASR \cite{ref28_DahlYDA12,ref29_KunzeKKKJS17} communities have attempted to tackle this issue by incorporating knowledge from pre-trained models trained on diverse set of large data and share commonalities across different tasks. The pre-trained acoustic and linguistic models can be finetuned for the downstream SLU tasks, which is proven to be an effective paradigm \cite{denisov2020pretrained, ref14_bidisha}. The study presented in~\cite{ref1_lugosch_speech_2019} explores acoustic feature extraction using a pre-trained ASR model from spoken utterances, which are further used for intent classification~\cite{ref14_bidisha}.

The neural transformers have shown encouraging performance on several benchmarks in the field of NLP~\cite{attention_VaswaniSPUJGKP17,ref_20_transformer_library_wolf-etal-2020-transformers}. However, their capabilities have not been fully investigated in the field of SLU. In \cite{ref5_dong_speech-transformer_2018}, transformer network has been successfully used for speech recognition. In \cite{ref3_radfar_end--end_2020}, the authors leverage the transformer architecture for end-to-end SLU. The self-attention mechanism in the transformer enables the extraction of the semantic context from a spoken utterance. Additionally, the self-attention allows to compute the correlation in each sub-layer between all pairs of time steps~\cite{attention_VaswaniSPUJGKP17}. This makes the transformer based SLU architecture more efficient in capturing longer sequence information than the recurrent neural network (RNN) based models~\cite{ref3_radfar_end--end_2020}. 



A well-trained model captures meaningful knowledge or information for a specific task. 
The knowledge distillation approach aims to distill the learning capacity of a larger deep neural network (teacher model) to a smaller network (student model) \cite{ref4_gou_knowledge_2021,HintonVD15}. It has shown efficacy in cross-modal scenarios, where the teacher model is trained on one modality and the knowledge is transferred to another modality~\cite{zhou2020unified,chuang2019speechbert}. Previous works in intent classification demonstrate the usefulness of knowledge distillation, which transfers the knowledge of linguistic embeddings extracted from a teacher model (language model), such as BERT to a student model (speech model)~\cite{huang2020leveraging,denisov2020pretrained,ref14_bidisha,futami2020distilling}.  

\begin{figure*}[!t]
\vspace{-0.2cm}
    \centering
      \includegraphics[trim=0cm 0cm 0cm 0.1cm, clip,scale=0.34]{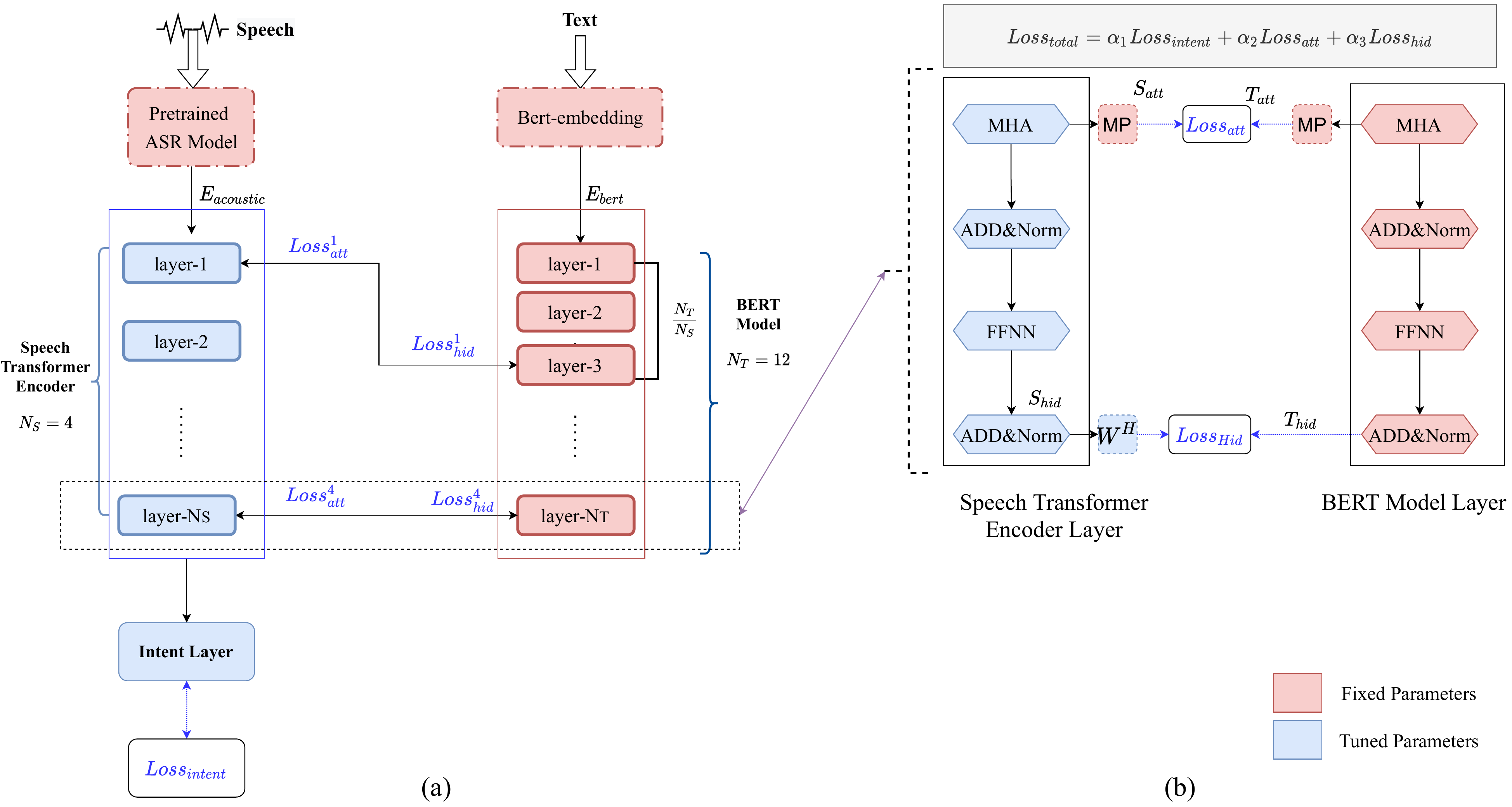}
    \vspace{-0.1cm}
    \caption{The proposed STD model: (a) overall architecture and different loss computation, (b) knowledge distillation losses (i.e., computation of $Loss_{att}$ and $Loss_{hid}$) between two transformer layers of student and teacher model expanded from dotted box in (a). }
    \label{fig:overview}
\end{figure*}


We observe that the attention weights capture rich linguistic information in the BERT model \cite{ref19_BERTanalysis}. The linguistic knowledge captured in the BERT model includes syntax and co-reference information, which are crucial for intent classification task~\cite{ref6_jiao_tinybert_2020}. In ~\cite{ref6_jiao_tinybert_2020}, an effective transformer-to-transformer knowledge distillation method has been proposed, where the knowledge encoded in a large teacher BERT model is transferred to a small student TinyBERT model. 
This specific type of distillation~\cite{ref6_jiao_tinybert_2020} consists of attention based and hidden state based distillation, which are performed from one layer of the teacher model to corresponding layer of the student model. Motivated from these, we propose the attention and hidden state based distillation from the teacher BERT model to a student transformer model. 

In this work, we are inspired by the success of the transformer based speech models for speech recognition tasks and the BERT model for various NLP tasks. As we know human interpretation of conversation or spoken utterance does not equally rely on all words in the utterance. Some words in a spoken utterance are more important than others to infer the meaning of the utterance. The attention mechanism in a speech transformer model allows us to look at the most relevant frames of speech embeddings based on speech-specific features. Similarly, transformer based language model pays more attention to the relevant words in a sentence based on linguistic knowledge. We would like to explore the transfer learning method, where information from both attention and hidden layers of the BERT model can be distilled to the speech transformer model. To achieve this, we propose a multilevel transformer based teacher-student model, where the knowledge distillation on both attention and hidden layers is performed for the intent classification task. We also use a pre-trained ASR model to extract speech embeddings from the spoken utterances. The proposed architecture reduces the resource requirement of transformer based SLU as we are employing both pre-trained ASR model and BERT model. Furthermore, the proposed transformer layer distillation leverages information from both linguistic and acoustic aspects.

\vspace{-0.2cm}
\section{Speech Transformer Distillation (STD) Framework}
\label{sec:proposed}
In this section, we discuss the proposed knowledge distillation approach, which we refer to as {\it speech transformer distillation} (STD) framework. Figure \ref{fig:overview} shows the proposed STD framework for intent classification. It mainly consists of five components, namely, pre-trained ASR model, speech-transformer encoder as student model, BERT model as teacher model, transformer-layers distillation and intent layer. Here, the parameters of the pre-trained ASR model, BERT-embedding extraction and BERT model are kept fixed, which do not change during the training process. The parameters of the speech transformer encoder and intent layer are updated during the training process.

Initially, we extract the acoustic embeddings from the pre-trained ASR model as described in \cite{ref1_lugosch_speech_2019} and BERT embeddings using the BertTokenizer package~\footnote{https://github.com/huggingface/transformers}. Then the acoustic embeddings ($E_{\text{acoustic}}$) and BERT embeddings ($E_{\text{bert}}$) are fed into the student and teacher models, respectively. Both the student and teacher networks compose of identical transformer based model structure. We perform transfer learning from one layer of the teacher model to the corresponding layer of the student model to minimize the distance based on two aspects, namely, attention matrices and hidden state vectors. The framework is optimized by three losses, namely, intent loss from ground-truth label, attention loss from the attention based distillation and hidden loss from the hidden state based distillation. Further details of each component of the proposed STD framework are as follows.
\vspace{-0.2cm}
\subsection{Pre-trained ASR Model}
\label{subsec:2.1-pretrainedASR}

We follow the same strategy to use a pre-trained ASR model as described in \cite{ref1_lugosch_speech_2019} and a 256-dimensional hidden representation is obtained from speech. 
These hidden representations are used as acoustic embeddings ($E_{\text{acoustic}}$) and applied as input to the student speech transformer model. More details about training strategy of the pre-trained ASR model can be referred from ~\cite{ref1_lugosch_speech_2019}. We freeze this pre-trained ASR model and don't update the parameters while using for downstream transformer model.
\vspace{-0.2cm}
\subsection{Speech Transformer Encoder}
\label{subsec:2.2-speechTransformer}
The proposed SLU encoder structure is based on the neural transformer network similar as~\cite{ref18_ZhaoLWL19}. The transformer architecture consists of three sub-modules: embedding, encoder and decoder. In the STD framework, we are using only the speech transformer encoder component. The encoder contains  $N_{\text{S}}$ identical transformer layers, and each layer has two sub-layers: multi-head attention (MHA) and position-wise feed-forward network (FFN) as shown in Figure~\ref{fig:overview} (a). This student speech transformer encoder model produces the attention matrices as $S_{att}$,  which are output from the MHA structures, and hidden state vectors as $S_{hid}$.

\subsection{BERT Embedding and Model}
\label{subsec:2.3-bert}
To take the benefit from the large unlabeled data and derive better representation of language model pre-training techniques are used in BERT model \cite{ref27_bert_paper}. The BERT model consists of multiple bidirectional transformer layers. In Figure \ref{fig:overview} (a) we show the BERT teacher model and the underlying structure of each layer is shown in Figure \ref{fig:overview} (b).  For each text utterance, the tokens obtain from the BertTokenizer are converted into embeddings $E_{bert}$, which are given as input to the BERT model. In different layers of the BERT model, we produce the attention matrices as $T_{att}$,  which are output from the MHA structures, and the hidden states as $T_{hid}$.  



\subsection{Transformer-layers Distillation}
\label{subsec:2.4-transformerDistillation}

In order to leverage information from linguistically rich informative BERT model into a speech model, we employ transformer-to-transformer distillation. This comprises of  the layer-by-layer attention based distillation and hidden state based distillation as shown in Figure \ref{fig:overview}. Figure \ref{fig:overview}(b) shows that the knowledge distillation is performed between each pair of transformer layers of the teacher and student models. 
 
Each distillation procession includes attention based distillation and hidden state based distillation. There are $N_{T}=12$ transformer layers in the BERT model, while we set $N_{S}=4$  layers in the student speech transformer encoder model. To perform a layer-by-layer distillation, we combine $N_{T}/N_{S}$ layers into a block, which results in 4 blocks in the BERT teacher model. Then, the last layer in each block is chosen for distillation with each layer in the speech transformer encoder structure. Figure \ref{fig:overview}(b) shows the transformer layer-by-layer distillation process corresponding to a pair of teacher-student layers. 

In order to perform the attention based distillation, we minimize mean square error (MSE) distance between the attention matrices of student and teacher models. We refer to this as $Loss_{att}$, which is computed as follows.
\begin{equation}
    Loss_{att} = \text{MSE}_{loss}(S_{att},T_{att}),
    \label{eq:loss_att}
\end{equation}
where, $MSE_{loss}$ indicates the mean squared error loss function. The number of attention matrices are dependent on the number of heads used in the student and teacher models. As shown in Figure~\ref{fig:overview}(b), we apply mean pooling (MP) to derive average of attention matrices over all attention heads (size $L \times L$), which are $S_{att}$ and $T_{att}$ for student and teacher model, respectively. $L$ is the sequence length.


Next, for the hidden state based distillation, we first derive the hidden state vectors of the transformer layers corresponding to the teacher and student models, which are $T_{hid}$ and $S_{hid}$, respectively. These are used to distill the knowledge from hidden state vectors as follows.
\begin{equation}
    Loss_{hid} = \text{MSE}_{loss}(S_{hid} W^H,T_{hid}),
    \label{eq:loss_hid}
\end{equation}
where, $S_{hid}\in\mathbb{R}^{L\times S_{dmodel}}$, $T_{hid}\in\mathbb{R}^{L\times T_{dmodel}}$.  Here we use a learnable linear transformation $W^{H}\in\mathbb{R}^{S_{dmodel}\times T_{dmodel}}$, which converts hidden states of student network into same dimension as the teacher network’s states.

\subsection{Intent Layer}
\label{subsec:2.5-IClayer}
The intent layer is responsible for the final intent classification after the knowledge distillation process. Here, we use a linear layer for intent classification.  The intent classification loss is referred to as $Loss_{intent}$, which is computed by cross-entropy calculation. 
The intent loss is calculated as follows:
\begin{equation}    
    Loss_{\text{intent}} = \text{CrossEntropy}(Intent_{\text{pred}} ,y),\\
    \label{eq:loss_intent}
\end{equation}

\noindent where, $Intent_{pred}$ is the prediction from intent layer, and  $y$ is the ground truth label. 


Next, we combine $Loss_{intent}$ with the two distillation losses ($Loss_{att}$ and $Loss_{hid}$) using the weights to make them in the similar scale. The total loss ($Loss_{total}$) is further used to back-propagate the whole framework.
\begin{equation}\label{total_loss}
    Loss_{total} = \alpha_1 Loss_{intent} + \alpha_2 Loss_{att} + \alpha_3 Loss_{hid},
\end{equation}
where, $\alpha_1$, $\alpha_2$, and $\alpha_3$ are the weights associated with three losses to tune the proportion of each loss in the total loss for better back-propagation.
\section{Experimental Details}
\label{sec:exp_details}
In this section, we describe the database used  and experimental details to validate the proposed STD framework for intent classification using speech. We have released the experimental set up of this work in the following link ~\footnote{https://github.com/Jiang-Yidi/TransformerDistillation-SLU}.
\subsection{Dataset}
\label{subsec:exp_details}
We evaluate the proposed STD framework on two well-known SLU datasets, namely, Fluent speech commands (FSC)~\cite{ref1_lugosch_speech_2019} and ATIS~\cite{ref15_ATIS}. Details of the datasets are presented in Table \ref{tab:table1-database}.


\begin{table}[!htbp]
  \centering
    \caption{Details of Fluent speech commands (FSC) and ATIS databases used.}
    \label{tab:table1-database}
    \resizebox{0.45\textwidth}{!}{
    \renewcommand\arraystretch{1.4}
  \begin{tabular}{|c|c|c|c|c|}
   \hline
    \multirow{2}*{\diagbox[width=6cm]{Specification}{Database}}&\multicolumn{2}{c|}{FSC}&\multicolumn{2}{c|}{ATIS}\cr\cline{2-5}
    &Train&Test&Train&Test\cr
    \hline
    \hline
     No. of utterances&23132&3793&5253&580\\
    \hline
    Total Duration (sec) &882.95&154.58&569.90&61.66\\
    \hline
    Avg duration of utterance (sec) &2.29&2.45&6.51&6.37\\
    \hline
    No. of intents&31&31&15&15\\
    \hline
  \end{tabular}
  }
\vspace{-0.3cm}
 \end{table}

\subsection{Baseline Systems}
\label{subsec:baseline}
To show the comparative performance of the proposed STD framework, we develop two baseline systems. In the first baseline ({\bf Baseline-1}), we initially extract the acoustic embedding vector ($E_{\text{acoustics}}$) for each spoken utterance using the pre-trained ASR model as discussed in Section~\ref{subsec:2.1-pretrainedASR}. The 256-dimensional $E_{\text{acoustics}}$ vector is further modeled using a RNN layer, which is a gated recurrent unit (GRU), followed by maxpooling, and linear layer for intent classification as discussed in ~\cite{ref1_lugosch_speech_2019}.

The second baseline ({\bf Baseline-2}) is the transformer based end-to-end SLU model~\cite{ref3_radfar_end--end_2020}, where, we first extract the 256-dimensional $E_{\text{acoustics}}$ from the pre-trained ASR model as in Section~\ref{subsec:2.1-pretrainedASR}. The $E_{\text{acoustics}}$ vector is passed through the speech transformer encoder as discussed in Section~\ref{subsec:2.2-speechTransformer} followed by the intent layer discussed in Section~\ref{subsec:2.5-IClayer}. The Baseline-2 system demonstrates the impact of the proposed transfer learning process when we compare with the proposed STD framework.

In addition to the above mentioned baselines, we also consider the original text to obtain the performance with only text based NLU model using BERT ({\bf BERT model}). 

\subsection{Proposed System}

In the proposed STD framework, we first extract $E_{acoustic}$ from each spoken utterance using the pre-trained ASR model and $E_{bert}$ from the corresponding text using the BERT embedding module. We pass $E_{acoustic}$ and $E_{bert}$ to the student and teacher models, respectively. 
For the teacher model, we use the \verb|bert-base-uncased| model without fine-tuning to capture the linguistic information. The teacher model consists of 12 transformer layers ($N_T=12$), each of which comprises of 12 attention heads ($T_{nhead}=12$) and 768 hidden units ($T_{dmodel}=768$) \cite{ref27_bert_paper}. For the student model, 8 no. of heads ($S_{nhead}=8$), 512 hidden units ($S_{dmodel}=512$) and 4 no. of layers ($N_S=4$) are set based on experimental evidence.

During training, two transformer-layer distillation losses as shown in Eq (\ref{eq:loss_att}), Eq (\ref{eq:loss_hid}), and intent loss in Eq (\ref{eq:loss_intent}) are used to back-propagate the speech transformer encoder and intent layers simultaneously. The values of $\alpha_1$, $\alpha_2$ and $\alpha_3$ are set as 0.625, 0.125 and 0.250, respectively to calculate the total loss.

To overcome the over-fitting while calculating the intent loss as per Eq (\ref{eq:loss_intent}), we use label smoothing as a regularization technique. We set the smoothing parameter as 0.1 during loss calculation as given in \footnote{https://github.com/kaituoxu/Speech-Transformer/blob/master/src/transformer/loss.py}. The optimization technique used for training the STD framework is Adam optimizer with $\beta_1=0.9$, $\beta_2=0.98$, $\epsilon=10^{-9}$ and learning rate scheduling is followed as mentioned in \cite{ref18_ZhaoLWL19}. The entire STD framework is implemented using the PyTorch library \footnote{https://pytorch.org/}. 

To investigate the effectiveness of individual distillation process, we develop two versions of the proposed STD framework. Firstly, we use only hidden state based distillation ({\bf Proposed: STD (hidden)}) and secondly, we include both attention and hidden state based distillation in the proposed STD framework ({\bf Proposed: STD}).

In order to analyze the impact of noise on the Proposed:STD model (with knowledge distillation) compared to the Baseline-2 (without knowledge distillation), we evaluate the performance of both the systems under acoustically degraded condition. We add babble noise from NOISEX-92 database\footnote{http://mi.eng.cam.ac.uk/comp.speech/Section1/Data/noisex.html} at different levels (15 dB, 10 dB, 5 dB, 0 dB) of signal-to-noise ratio (SNR) to the speech signals corresponding to the test data of both ATIS and FSC databases~\cite{sharma2016sonority,sharma2016speech}. The degraded test speech signals are passed through the Baseline-2 and proposed STD systems to test the respective performances.

\section{Experimental Results}
\label{sec:res}
In Table \ref{tab:tab2-result}, we present the intent classification accuracy for the BERT, baseline and the proposed models. We observe that the proposed STD framework improves the intent classification performance as compared to the Baseline-1 and Baseline-2 (no distillation). While using the only hidden state based distillation (Proposed:STD(hidden)), the result indicates that only hidden state vector contributes to significant performance improvement as compared to Baseline-2 (no distillation). Further, using both attention and hidden state based knowledge distillation (Proposed:STD), we achieve a reasonably good performance in the intent classification task for both FSC and ATIS databases.

Table \ref{tab:table3-result-noise} shows that the performance of the proposed STD framework is better than the Baseline-2 in presence of noise with different SNR levels. The intent classification accuracy degrades as SNR level decreases. Specifically, we observe more degradation in performance with noise for the FSC database than the ATIS database. We note that the Baseline-2 refers to the conventional speech transformer SLU model without distillation. This indicates the stability of the proposed STD framework against the acoustic degradation.

\begin{table}[!htbp]
    \centering
    \caption{Intent classification accuracy (\%) of baseline methods and proposed STD model for Fluent speech commands (FSC) and ATIS databases.}
    \label{tab:tab2-result}
    \scalebox{0.88}{
    \begin{tabular}{|c|c|c|}
    \hline
    Model& FSC& ATIS\\
    \hline
    \hline
    BERT model  & 100& 94.48\\
    (using original text) & &\\
    \hline
    Baseline-1 \cite{ref1_lugosch_speech_2019} & 98.80 & 85.43\\
    \hline
    Baseline-2 & 98.87& 86.55\\
    \hline
Proposed:     STD (hidden) & \textbf{99.00} & \textbf{88.44}\\
    \hline
    Proposed: STD & \textbf{99.10} & \textbf{88.79}\\
    \hline
    \end{tabular}}
\end{table}

\vspace{-0.2cm}
\begin{table}[ht]
\centerline{
\ninept \caption{\label{tab:table3-result-noise} {Intent classification performance of Baseline-2 and proposed STD model for Fluent speech commands (FSC) and ATIS databases under acoustically degraded condition.}}
\renewcommand{\arraystretch}{1.2}
\scalebox{0.8}{
\begin{tabular}{|c c c c c c|}
\hline
\multicolumn{2}{|c}{Noise level (dB) $\rightarrow$}  & \multicolumn{1}{|c|}{\multirow{2}{*}{15}} & \multicolumn{1}{|c|}{\multirow{2}{*}{10}} & \multicolumn{1}{|c|}{\multirow{2}{*}{5}} & \multicolumn{1}{|c|}{\multirow{2}{*}{0}}\\
\cline{1-2}
\multicolumn{1}{|c}{Database $\downarrow$}  & \multicolumn{1}{|c}{Model $\downarrow$}  & \multicolumn{1}{|c|}{} & \multicolumn{1}{|c|}{} & \multicolumn{1}{|c|}{} & \multicolumn{1}{|c|}{}\\
\hline
\hline
\multicolumn{1}{|c}{\multirow{2}{*}{FSC}}  & \multicolumn{1}{|c|}{Baseline-2}  & \multicolumn{1}{|c|}{93.20} & \multicolumn{1}{|c|}{74.80} & \multicolumn{1}{|c|}{45.90} & \multicolumn{1}{|c|}{14.90}\\
\cline{2-6}
\multicolumn{1}{|c|}{}  & \multicolumn{1}{|c}{Proposed:STD}  & \multicolumn{1}{|c|}{95.60} & \multicolumn{1}{|c|}{78.90} & \multicolumn{1}{|c|}{48.20} & \multicolumn{1}{|c|}{17.60}\\
\hline
\multicolumn{1}{|c}{\multirow{2}{*}{ATIS}}  & \multicolumn{1}{|c}{Baseline-2}  & \multicolumn{1}{|c|}{82.41} & \multicolumn{1}{|c|}{81.03} & \multicolumn{1}{|c|}{76.90} & \multicolumn{1}{|c|}{71.03}\\
\cline{2-6}
\multicolumn{1}{|c}{}  & \multicolumn{1}{|c}{Proposed:STD}  & \multicolumn{1}{|c|}{86.90} & \multicolumn{1}{|c|}{85.86} & \multicolumn{1}{|c|}{81.21} & \multicolumn{1}{|c|}{72.59}\\
\hline
\end{tabular}}
}
\end{table}

\vspace{-0.2cm}
\section{Conclusions}
\label{sec:summary}
In this paper, we introduce a knowledge distillation approach from BERT transformer to speech transformer for intent classification using speech. 
We demonstrate that the transformer based distillation helps to improve speech transformer model-based SLU for intent classification. In particular, we perform layer-by-layer knowledge distillation with respect to two components of the transformer model, namely, hidden state vectors and attention matrices. The experimental results on FSC and ATIS databases show improved accuracy after incorporation of transformer based knowledge distillation. Thus, we observe that the knowledge distillation can learn from rich informative BERT model and improve the intent classification performance using only speech. The experiments under noisy condition confirm the better robustness of the proposed architecture. In future, we would like to explore the possible temporal alignment across speech and text for intent classification. 

\section{Acknowledgements}
This research is supported by the Agency for Science, Technology and Research (A*STAR) under its AME Programmatic Funding Scheme (Project No. A18A2b0046) and Science and Engineering Research Council, Agency of Science, Technology and Research, Singapore, through the National Robotics Program under Grant No. 192 25 00054.

\newpage
\footnotesize
\bibliographystyle{ieeetr}  
\bibliography{IntentClassification.bib} 

\begin{thebibliography}{10}

\bibitem{ref7_deepspeech2}
D.~Amodei {\em et~al.}, ``Deep speech 2 : End-to-end speech recognition in
  {English} and mandarin,'' in {\em {ICML}}, vol.~48, pp.~173--182, 2016.

\bibitem{ref8_ChanJLV16}
W.~Chan, N.~Jaitly, Q.~V. Le, and O.~Vinyals, ``Listen, attend and spell: A
  neural network for large vocabulary conversational speech recognition,'' in
  {\em {ICASSP}}, pp.~4960--4964, 2016.

\bibitem{ref9_neuralasr}
H.~Soltau, H.~Liao, and H.~Sak, ``Neural speech recognizer: Acoustic-to-word
  {LSTM} model for large vocabulary speech recognition,'' in {\em Interspeech},
  pp.~3707--3711, 2017.

\bibitem{ref10_OordDZSVGKSK16}
A.~van~den Oord, S.~Dieleman, H.~Zen, K.~Simonyan, O.~Vinyals, A.~Graves,
  N.~Kalchbrenner, A.~W. Senior, and K.~Kavukcuoglu, ``Wavenet: {A} generative
  model for raw audio,'' in {\em {ISCA} Speech Synthesis Workshop}, p.~125,
  2016.

\bibitem{ref11_SutskeverVL14}
I.~Sutskever, O.~Vinyals, and Q.~V. Le, ``Sequence to sequence learning with
  neural networks,'' in {\em Annual Conference on Neural Information Processing
  Systems}, pp.~3104--3112, 2014.

\bibitem{ref12_BahdanauCB14}
D.~Bahdanau, K.~Cho, and Y.~Bengio, ``Neural machine translation by jointly
  learning to align and translate,'' in {\em 3rd International Conference on
  Learning Representations, {ICLR}}, 2015.

\bibitem{ref13_CNNseq2ses}
J.~Gehring, M.~Auli, D.~Grangier, D.~Yarats, and Y.~N. Dauphin, ``Convolutional
  sequence to sequence learning,'' in {\em International Conference on Machine
  Learning, {ICML}}, vol.~70, pp.~1243--1252, 2017.

\bibitem{serdyuk2018towards}
D.~Serdyuk, Y.~Wang, C.~Fuegen, A.~Kumar, B.~Liu, and Y.~Bengio, ``Towards
  end-to-end spoken language understanding,'' in {\em 2018 IEEE International
  Conference on Acoustics, Speech and Signal Processing (ICASSP)},
  pp.~5754--5758, IEEE, 2018.

\bibitem{price2020improved}
R.~Price, M.~Mehrabani, and S.~Bangalore, ``Improved end-to-end spoken
  utterance classification with a self-attention acoustic classifier,'' in {\em
  ICASSP 2020-2020 IEEE International Conference on Acoustics, Speech and
  Signal Processing (ICASSP)}, pp.~8504--8508, IEEE, 2020.

\bibitem{haghani2018audio}
P.~Haghani, A.~Narayanan, M.~Bacchiani, G.~Chuang, N.~Gaur, P.~Moreno,
  R.~Prabhavalkar, Z.~Qu, and A.~Waters, ``From audio to semantics: Approaches
  to end-to-end spoken language understanding,'' in {\em IEEE Spoken Language
  Technology Workshop (SLT)}, pp.~720--726, 2018.

\bibitem{ref1_lugosch_speech_2019}
L.~Lugosch, M.~Ravanelli, P.~Ignoto, V.~S. Tomar, and Y.~Bengio, ``Speech model
  pre-training for end-to-end spoken language understanding,'' in {\em
  Interspeech}, pp.~814--818, Sept. 2019.

\bibitem{ref21_YosinskiCBL14}
J.~Yosinski, J.~Clune, Y.~Bengio, and H.~Lipson, ``How transferable are
  features in deep neural networks?,'' in {\em Advances in Neural Information
  Processing Systems 27: Annual Conference on Neural Information Processing
  Systems}, pp.~3320--3328, 2014.

\bibitem{ref22_SimonyanZ14a}
K.~Simonyan and A.~Zisserman, ``Very deep convolutional networks for
  large-scale image recognition,'' in {\em International Conference on Learning
  Representations, {ICLR}}, 2015.

\bibitem{zhou2020unified}
L.~Zhou, H.~Palangi, L.~Zhang, H.~Hu, J.~Corso, and J.~Gao, ``Unified
  vision-language pre-training for image captioning and vqa,'' in {\em
  Proceedings of the AAAI Conference on Artificial Intelligence}, vol.~34,
  pp.~13041--13049, 2020.

\bibitem{ref23_DaiL15}
A.~M. Dai and Q.~V. Le, ``Semi-supervised sequence learning,'' in {\em Advances
  in Neural Information Processing Systems 28: Annual Conference on Neural
  Information Processing Systems}, pp.~3079--3087, 2015.

\bibitem{ref24_RuderH18}
J.~Howard and S.~Ruder, ``Universal language model fine-tuning for text
  classification,'' in {\em ACL}, pp.~328--339, Association for Computational
  Linguistics, 2018.

\bibitem{ref25_radford2018improving}
A.~Radford, K.~Narasimhan, T.~Salimans, and I.~Sutskever, ``Improving language
  understanding by generative pre-training,'' in {\em Technical report,
  OpenAI}, 2018.

\bibitem{ref26_SanhWR19}
V.~Sanh, T.~Wolf, and S.~Ruder, ``A hierarchical multi-task approach for
  learning embeddings from semantic tasks,'' in {\em AAAI}, pp.~6949--6956,
  2019.

\bibitem{ref27_bert_paper}
J.~Devlin, M.~Chang, K.~Lee, and K.~Toutanova, ``{BERT:} pre-training of deep
  bidirectional transformers for language understanding,'' in {\em
  {NAACL-HLT}}, pp.~4171--4186, 2019.

\bibitem{ref28_DahlYDA12}
G.~E. Dahl, D.~Yu, L.~Deng, and A.~Acero, ``Context-dependent pre-trained deep
  neural networks for large-vocabulary speech recognition,'' {\em {IEEE} Trans.
  Speech Audio Process.}, vol.~20, no.~1, pp.~30--42, 2012.

\bibitem{ref29_KunzeKKKJS17}
J.~Kunze, L.~Kirsch, I.~Kurenkov, A.~Krug, J.~Johannsmeier, and S.~Stober,
  ``Transfer learning for speech recognition on a budget,'' in {\em Proc. of
  the 2nd Workshop on Representation Learning for NLP, Rep4NLP@ACL},
  pp.~168--177, Association for Computational Linguistics, 2017.

\bibitem{denisov2020pretrained}
P.~Denisov and N.~T. Vu, ``Pretrained semantic speech embeddings for end-to-end
  spoken language understanding via cross-modal teacher-student learning,'' in
  {\em Proc. Interspeech}, pp.~881--885, 2020.

\bibitem{ref14_bidisha}
B.~Sharma, M.~C. Madhavi, and H.~Li, ``Leveraging acoustic and linguistic
  embeddings from pretrained speech and language models for intent
  classification,'' in {\em Accepted in ICASSP 2021}, 2021.

\bibitem{attention_VaswaniSPUJGKP17}
A.~Vaswani, N.~Shazeer, N.~Parmar, J.~Uszkoreit, L.~Jones, A.~N. Gomez,
  L.~Kaiser, and I.~Polosukhin, ``Attention is all you need,'' in {\em Advances
  in Neural Information Processing Systems 30: Annual Conference on Neural
  Information Processing Systems}, pp.~5998--6008, 2017.

\bibitem{ref_20_transformer_library_wolf-etal-2020-transformers}
T.~Wolf {\em et~al.}, ``Transformers: State-of-the-art natural language
  processing,'' in {\em Proceedings of the 2020 Conference on Empirical Methods
  in Natural Language Processing: System Demonstrations}, pp.~38--45,
  Association for Computational Linguistics, 2020.

\bibitem{ref5_dong_speech-transformer_2018}
L.~Dong, S.~Xu, and B.~Xu, ``Speech-transformer: A no-recurrence
  sequence-to-sequence model for speech recognition,'' in {\em {ICASSP}},
  (Calgary, AB), pp.~5884--5888, Apr. 2018.

\bibitem{ref3_radfar_end--end_2020}
M.~Radfar, A.~Mouchtaris, and S.~Kunzmann, ``End-to-end neural transformer
  based spoken language understanding,'' in {\em Interspeech}, pp.~866--870,
  ISCA, 2020.

\bibitem{ref4_gou_knowledge_2021}
J.~Gou, B.~Yu, S.~J. Maybank, and D.~Tao, ``Knowledge distillation: A survey,''
  {\em arXiv:2006.05525 [cs, stat]}, mar 2021.
\newblock arXiv: 2006.05525.

\bibitem{HintonVD15}
G.~Hinton, O.~Vinyals, and J.~Dean, ``Distilling the knowledge in a neural
  network,'' in {\em NIPS Deep Learning and Representation Learning Workshop},
  2015.

\bibitem{chuang2019speechbert}
Y.-S. Chuang, C.-L. Liu, H.~yi~Lee, and L.~shan Lee, ``{SpeechBERT: An
  Audio-and-Text Jointly Learned Language Model for End-to-End Spoken Question
  Answering},'' in {\em Proc. Interspeech 2020}, pp.~4168--4172, 2020.

\bibitem{huang2020leveraging}
Y.~Huang, H.-K. Kuo, S.~Thomas, Z.~Kons, K.~Audhkhasi, B.~Kingsbury, R.~Hoory,
  and M.~Picheny, ``Leveraging unpaired text data for training end-to-end
  speech-to-intent systems,'' in {\em ICASSP}, pp.~5754--5758, 2020.

\bibitem{futami2020distilling}
H.~Futami, H.~Inaguma, S.~Ueno, M.~Mimura, S.~Sakai, and T.~Kawahara,
  ``Distilling the knowledge of bert for sequence-to-sequence asr,'' {\em arXiv
  preprint arXiv:2008.03822}, 2020.

\bibitem{ref19_BERTanalysis}
K.~Clark, U.~Khandelwal, O.~Levy, and C.~D. Manning, ``What does {BERT} look
  at? an analysis of bert's attention,'' {\em CoRR}, vol.~abs/1906.04341, 2019.

\bibitem{ref6_jiao_tinybert_2020}
X.~Jiao, Y.~Yin, L.~Shang, X.~Jiang, X.~Chen, L.~Li, F.~Wang, and Q.~Liu,
  ``{TinyBERT}: Distilling {BERT} for natural language understanding,'' in {\em
  Findings of the Association for Computational Linguistics: {EMNLP}},
  pp.~4163--4174, 2020.

\bibitem{ref18_ZhaoLWL19}
Y.~Zhao, J.~Li, X.~Wang, and Y.~Li, ``The speech transformer for large-scale
  mandarin chinese speech recognition,'' in {\em {ICASSP}}, pp.~7095--7099,
  {IEEE}, 2019.

\bibitem{ref15_ATIS}
P.~J. Price, ``Evaluation of spoken language systems: the {ATIS} domain,'' in
  {\em Speech and Natural Language: Proceedings of a Workshop}, Morgan
  Kaufmann, 1990.

\bibitem{sharma2016sonority}
B.~Sharma and S.~R.~M. Prasanna, ``Sonority measurement using system, source,
  and suprasegmental information,'' {\em IEEE/ACM Transactions on Audio,
  Speech, and Language Processing}, vol.~25, no.~3, pp.~505--518, 2016.

\bibitem{sharma2016speech}
B.~Sharma and S.~R.~M. Prasanna, ``Speech synthesis in noisy environment by
  enhancing strength of excitation and formant prominence.,'' in {\em
  INTERSPEECH}, pp.~131--135, 2016.

\end{thebibliography}

\end{document}